\documentclass{llncs}
\usepackage[utf8x]{inputenc}

\usepackage{amsmath}
\usepackage{amsfonts}

\usepackage{graphicx}
\usepackage{tikz}
\usepackage{paralist}

\usepackage{caption}
\usepackage{subcaption}
\usepackage{placeins}

\begin{document}

\title{Key point selection and clustering of swimmer coordination through Sparse Fisher-EM}
\titlerunning{Key point selection and clustering}  
%
\author{John Komar\inst{1} \and Romain Herault\inst{2} \and Ludovic Seifert\inst{1}}
\authorrunning{John Komar et al.} 
%
\tocauthor{John Komar, Romain Herault, Ludovic Seifert}
\institute{CETAPS EA-3832 Université de Rouen,\\
Boulevard Siegfried, 76821 Mont Saint Aignan, France\\
\email{firstname.lastname@univ-rouen.fr}
\and
LITIS EA-4108, INSA de Rouen,\\
Avenue de l'Université - BP 8,\\
76801 Saint-Étienne-du-Rouvray Cedex, France\\
\email{firstname.lastname@insa-rouen.fr}
\thanks{Authors would like to thank the \emph{Agence Nationale de la Recherche} (ANR) for its financial support to§ the project LeMOn (Learning with Multi-objective Optimization, ANR-11-JS02-10). }
}

\maketitle              

\begin{abstract} 
To answer the existence of optimal swimmer learning/teaching strategies,
this work introduces a two-level clustering in order to analyze temporal dynamics of motor learning in breaststroke swimming.
Each level have been performed through Sparse Fisher-EM, a unsupervised framework which can be applied efficiently on large and correlated datasets.
The induced sparsity selects key points of the coordination phase without any prior knowledge.
\end{abstract}

\section{Introduction}

The development of Dynamical Systems Theory \cite{kelso1995} in understanding motor learning has increased the interest of sports scientists in focusing on temporal dynamics of human motor behavior.
Broadly speaking, the investigation of motor learning traditionally implied the assessment of both a pre-learning behavior and a post-learning behavior \cite{muller2004}, but the deep understanding of the process of motor learning requires a continuous and long term assessment of the behavior rather than previous traditional discrete assessments.
Indeed, such a continuous assessment of behavioral data enables to investigate the nature of the learning process and might highlight the paramount role played by motor variability in optimizing learning \cite{muller2004}. 

From a theoretical point of view, motor learning is viewed as a process involving active exploration of a so-called perceptual-motor workspace which is learner dependent and defines all the motor possibilities available to him.
Few studies have already highlighted this exploratory behavior during learning a ski simulator task \cite{nourrit2003} or a soccer kicking task \cite{chow2008}.
These authors showed that learners exhibited different qualitative motor organizations during skill acquisition.
Nevertheless, these princeps studies mainly focused on a static analysis, defining the different behaviors exhibited during learning.
As a matter of fact, a major interest in the field of motor learning resides in the definition of different pathways of learning, namely different possible learning strategies \cite{gelfand1962}.
Such an interest in investigating the existence of different "routes of learning" needs to focus on a dynamical analysis, namely the analysis of the successions of different behaviors.
An unanswered question to date concerns the existence of optimal learning strategies (i.e. strategies that would appear more effective).
Thus, the discovery of optimal learning strategies could have a huge impact on the pedagogical approach of practitioners.

The article will describe at first the context of the research insisting on the way data have been collected,
what are the long-term expectations in sport science field and what are the short term locks in machine learning field.
Then we will give a brief view of the Fisher-EM algorithm \cite{bouv2012} which is an unsupervised learning method used in this work.
In the end, preliminary results of the data clustering will be analyzed.

\section{Context of the Research}

\subsection{Previous work}
In breaststroke swimming, achieving high performance requires a particular management of both arm and leg movements, in order to maximize propulsive effectiveness and optimize the glide and recovery times \cite{seifert2010}.
Therefore, expertise in breaststroke is defined by adopting a precise coordination pattern between arms and legs (i.e. a specific spatial and temporal relationship between elbow and knee oscillations).
Indeed, when knees are flexing, elbows should be fully extended (180°), whereas knees should be fully extended (180°) when elbows are flexing, in order to ensure a hydrodynamic position of the non-propulsive limbs when the first pair of limbs is actually propulsive \cite{seifert2005,tagaki2004}. 

Based on this context, the breaststroke swimming task was deemed as suitable in investigating the dynamics of learning, mainly as it implies at a macroscopic scale the acquisition of an expert arm-leg coordination that can be easily assessed.
however, the investigation of potential differences in learning strategies required a continuous movement assessment.
In that sense, the use of motion sensors allowed a fast, accurate and cycle per cycle movement assessment.

Previously, two analysis methods were used in the cycle per cycle study of motor learning. 
A previous study \cite{nourrit2003} highlighted the unstable character of the transition between novice and expert,
but not really an exploration as experimental setup assumes that novices left their initial behavior to adopt the expert one. 
Therefore, no search strategies were really investigated. 
In order to overcome this issue, \cite{chow2008} used a cluster analysis (Hierarchical Cluster Analysis) in their experiment
on football kicking and highlighted different behaviors used by each participant during learning to kick a ball.
The authors therefore linked these different behaviors to a search strategy.
However, the cluster analysis was performed individually and there was no comparison done between the learners (e.g. did they use identical behaviors?),
it implied only few participants (i.e. four learners), it was performed only with 120 kicks per learner (i.e. 10 kicks per session during 12 sessions)
and like the previous study of \cite{nourrit2003} it only defined the behavior from a static point of view (i.e. defining what behavior was adopted). 

\begin{figure}
\centering
\begin{tikzpicture}
\node[anchor=south west,inner sep=0] (fig) at (0,0) {\includegraphics[width=0.7\textwidth]{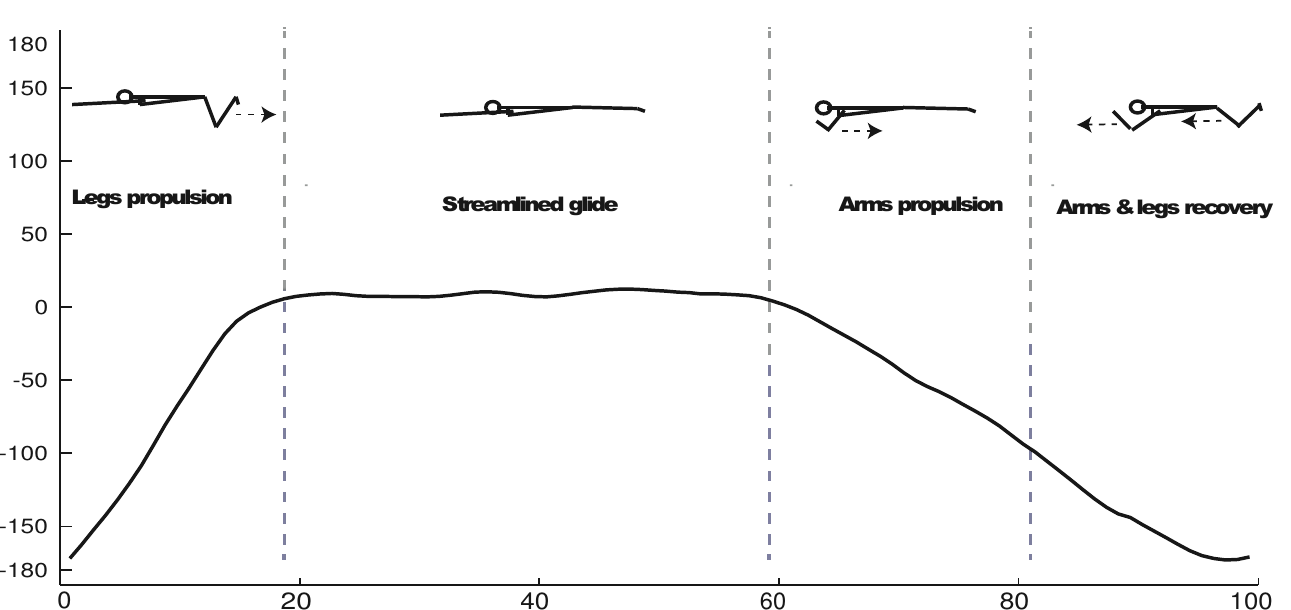}};
\path (fig.south)--+(0mm,-2mm) node {\tiny Percentage of cycle duration (\%)};
\path (fig.west)--+(-2mm,0mm) node[rotate=90] {\tiny Elbow-Knee relative phase (°)};
\end{tikzpicture}
\caption{A typical continuous relative phase between the knee and the elbow}
\label{fig:typcoord}
\end{figure}

\subsection{Data collection}

For this study, 26 novices were involved in 16 lessons of breaststroke swimming, with two sessions per week for a total duration of two months.
The general goal of learning for all the 26 swimmers was to increase the distance per stroke, while maintaining the speed stable.
Then the 26 learners were divided into four different groups, each group receiving a different instruction during the learning process:

\begin{inparaenum}[1)]
\item Control group (N=7): This group received only the general goal of learning, increase the distance per stroke
\item Analogy group (N=7): In addition to the general goal of learning, this group received a single additional instruction: "glide two seconds with your arms outstretched"
\item Pacer group (N=6): In addition to the general goal of learning, this group had to follow an auditory metronome trying to perform one cycle every single auditory signal.
The frequency of the metronome was decreased every two sessions, in order to promote a decrease in the stroke frequency of the learners that should lead to an increase in the distance per stroke
\item Prescription group (N=6): In addition to the general goal of learning, this group received multiple additional instructions: "keep your arms outstretched forward when you extend your legs; then glide with your arms and legs outstretched; then keep your legs outstretched when you flex your arms; recover both arms and legs together".
\end{inparaenum}
These different instructions were supposed to have a specific impact on the learning strategies of the learners.

Each learner performed 10 trials of 25-m swim during each session, with 1 x 25-m consisting approximatively in 8 recorded cycles
(one cycle correspond to the period between two successive maximal knee flexion).
During every learning session, all learners were equipped with small motion sensors on both arms and legs
(3-D gyroscopes, 3-D magnetometers, 3-D accelerometers) including a data logger and recording elbow and knee angles at a frequency of 200 Hz.

Following the literature in coordination dynamics \cite{kelso1995}, the coordination between elbow and knee was defined by the continuous relative phase between these two oscillators \cite{hamill2000},
considering elbows and knees as acting like individual pendulums \cite{seifert2010}.
A value of relative phase close to -180° or 180° defined an anti-phase relationship (i.e. opposite movements of knee and elbow)
while a value close to 0° defined an in-phase mode of coordination (i.e. identical movements of knee and elbow);
here, each cycle will be described by a time series of 100 normalized values of continuous relative phase between the knee and the elbow (Fig. \ref{fig:typcoord}).

To sum-up, we have recorded 4160 trials (26 swimmers $\times$ 16 sessions $\times$ 10 trials) and there is an average of 8 cycles per trials.
Thus, the dataset is composed by 33280 cycles,  each cycle is represented by 100 continuous relative phase samples.

\subsection{Study expectations}

From a sport sciences point of view, the specific aims of the study were twofold: 
\begin{inparaitem}
\item Assessing the dynamics of learning: In other words, the aim was to assess not only the different behaviors used during learning but also the transitions between these behaviors,
that is the potential search strategy exhibited by learners (e.g. they used preferably behavior n$^o$ 1 then n$^o$ 4, then n$^o$ 3 \ldots). 
\item Assessing the impact of different learning conditions on the dynamics of learning:
In other words, the aim was to investigate the possible existence of different behaviors exhibited by the learners regarding their learning condition,
as well as the possible existence of different search strategy exhibited by the different groups.
\end{inparaitem}

A last point in this experiment was the possibility to transfer the results of the analysis towards practical application or guidelines for teachers.
From a pedagogical point of view, it appeared difficult to teach novice swimmers by giving instruction on the arm-leg coordination during all the cycle and the definition of key points within the entire cycle reflects a paramount aspect for teaching.
Indeed, a strong literature in sports pedagogy highlights the role played by attentional focalization during motor learning, as a focalization on a key point of the swimming cycle may be highly beneficial in seeking to reorganize the entire arm-leg coordination \cite{Komar2013b}.
A third aim of this study was then to define highly discriminative key points within the swimming cycle and that might be the target of the instruction in order to orient the attention of learners.

From a machine learning point of view, there are two locks to tackle:
\begin{inparaenum}[1)]
\item Each cycle is described by 100 features which are highly correlated due to the fact that they are samples of the relative phase which is a continuous time signal. Nevertheless, we don't want to bias the study by preprocessing the data, a transformation like filters, wavelet transform or sample selection that will embedded our a priori knowledge. 
\item The number of cycles are not equal on all the trials, that is why a trial can not be directly described by a fixed number of features.
\end{inparaenum}

Those two problems were address by
\begin{inparaenum}[1)]
\item using a clustering by Fisher-EM~\cite{bouv2012} that also performs dimension reduction and features selection,
\item doing a two stage clustering: on cycles then on trials; a procedure similar to \emph{Bags of words} to have fixed size features on trial.
\end{inparaenum}

\section{Fisher-EM Algorithm}
A clustering can be derived from a mixture of Gaussians generative model.
A Gaussian, which is parameterized by a covariance matrix and a mean in the observation space, represents a cluster.
An observation is labeled according to its ownership (likelihood ratio) to each Gaussian.
Knowing the number of clusters, the  mixture and Gaussian parameters are learned from the observation data trough an Expectation-Maximization (EM) algorithm.

The Fisher-EM algorithm \cite{bouv2012} is based on the same principles but the mixture of Gaussians does not lie directly on the observation space but on a lower dimension latent space.
This latent space is chosen to maximize the Fisher criterion between clusters and thus be discriminative and its dimension is bounded by the number of clusters.
This reduction of dimension leads to more efficient computation on medium to large datasets (here 33280 examples by 100 features) as operations can be held in the smaller latent space.

\subsection{Generative Model}
We consider that the $n$ observations ${y_1, y_2, \ldots, y_n}$ are realizations of a random vector $Y\in\mathbb{R}^p$. We want to cluster these observations into $K$ groups.
For each observation $y_i$, a variable $z_i \in Z=\{1, \ldots, K\}$ indicates which cluster its belong to.
This clustering will be decided upon a generative model, namely a mixture of $K$ Gaussians which lies in a discriminative latent space $X\in\mathbb{R}^d$ where $d \leq K-1$.

This latent space is linked to the observation space through a linear transformation,
\begin{equation}
Y=UX+\epsilon\enspace, \label{eq:projection}
\end{equation}
\noindent where $U\in\mathbb{R}^{p\times d}$ and $U^tU=Id(d)$ where $Id(d)$ is the identity matrix of size $d$, i.e. $U$ is an orthogonal matrix and $\epsilon$ non-discriminative noise.

Let be $W=[U,V]\in\mathbb{R}^{p \times p}$ such that $W^tW=Id(p)$. $V$ is the orthogonal complement of $U$.
Thus, a projection $U^ty$  of an observation $y$ from space $Y$ of dimension $p$,
lies on the latent discriminative subspace $X$ of dimension $d$ and the projection $V^ty_i$  lies on the non-discriminative complement subspace of dimension $p-d$.

Conditionally to $Z=k$, random variables $X$ and $Y$ are assumed to be Gaussian,
$X_{|Z=k} \sim \mathcal{N}(\mu_k,\Sigma_k)\enspace,$ and
$Y_{|Z=k} \sim \mathcal{N}(m_k,S_k)\enspace, $  where $\mu_k\in\mathbb{R}^{d}$, $\Sigma_k\in\mathbb{R}^{d \times d}$, $m_k\in\mathbb{R}^{p}$ and $S_k\in\mathbb{R}^{p \times p}$.

With the help of equation~\ref{eq:projection}, we can deduce parameters of the distribution $Y_{|Z=k}$ in the observation space from the parameters of the distribution  $X_{|Z=k}$ in the latent space,
$m_k=U\mu_k$ and $S_k=U\Sigma_kU^t+\Psi\enspace,$ where  $\Psi\in\mathbb{R}^{p\times p}$ is the covariance matrix of $\epsilon$ which is assumed to follow a 0-centered Gaussian distribution.
To ensure that $\epsilon$ represents non-discriminative noise, we will impose that the covariance of $\epsilon$, $\Psi$, projected into the discriminative space is null,
i.e.  \mbox{$U\Psi U^t=\mathbf{0}(d)$},
and that $\Psi$ projected into the non-discriminative subspace is diagonal,
i.e. \mbox{$V\Psi V^t=\beta Id(p-d)$}.
Thus,
\begin{equation}
W^tS_kW=\left(\begin{array}{cc}
               \Sigma_k&\mathbf{0}\\
               \mathbf{0}&\beta_k Id(p-d)
              \end{array}\right)\enspace.
\end{equation}

All the Gaussian distributions are mixed together, the density of the generative model is given by
$f(y)=\sum_{k=1}^K \pi_k  \phi(y; m_k,S_k)$
where $\pi_k$ are mixing proportion and $m_k,S_k$ are deduced from $\{U,\beta,\mu_k,\Sigma_k\}$.

Finally, the model is parameterized by:
\begin{inparaitem}
\item $U$ the projection from discriminative subspace to observation space,
\item $\beta_k$ variance of $\epsilon$ in the non-discriminative subspace,
\item $\pi_k$ the mixing parameter,
\item and Gaussian parameter $\{\mu_k,\Sigma_k\}$,
\end{inparaitem}
\noindent where the 3 last parameters are repeated by the number of Gaussians.

Model variations, that lead to reduced numbers of parameters, can be achieved by enforcing shared covariances $\beta$ and/or $\Sigma$ between Gaussians,
diagonalization of the covariance $\Sigma$ without or with constant diagonal, and combination of these enforcements.

\subsection{Parameter estimation}

The iterative Expectation-Maximization (EM) algorithm can be extended  by a Fisher Step (\emph{F-Step}) in-between the \emph{E-Step} and the \emph{M-Step}  where  the latent discriminative subspace is computed  \cite{bouv2012}.
The Fisher criterion computed at the \emph{F-Step} is used as a stopping criterion. Convergences properties can be found in \cite{DBLP:journals/ma/BouveyronB12}.

\paragraph{E-Step} In this step, for each observation $i$, its posterior probability to each cluster $k$ is computed by
\begin{equation*}
o_{ik}\leftarrow \frac{\pi_k \phi(y_i,\hat{\theta}_k)}{\sum_{l=1}^K \pi_l \phi(y_i,\hat{\theta}_l)}\enspace,
\end{equation*}
\noindent where $\hat{\theta}_k=\{U,\beta,\mu_k,\Sigma_k\}$.
From these probabilities, each observation can be given to a cluster by $z_i=\underset{k}{\arg\max}\enspace o_{ik}$.

\paragraph{F-Step} The projection matrix $U$ is computed such that Fisher's criterion is maximized in the latent space,
\begin{equation*}
U\leftarrow \begin{array}{cc} \underset{U}{\arg\max}& trace\left(\left(U^tSU\right)^{-1} U^tS_BU\right)\\
             w.r.t. & U^tU=Id(d)
            \end{array} \enspace,
\end{equation*}
where $S$ is the variance of the whole dataset and
$S_B=\frac{1}{n}\sum_{k=1}^K n_k (m_k - \bar{y}) (m_k - \bar{y})^t$
\noindent where $n_k=\sum_i o_{ik}$ and $\bar{y}$ the mean of the dataset.

\paragraph{M-Step} Knowing the posterior probabilities $o_{ik}$ and the projection matrix $U$, we compute the new Gaussian parameters by maximizing the likelihood of the observations,

\begin{equation*}
\small
\hat{\pi}_k\leftarrow\frac{n_k}{n},\enspace
\hat{\mu}_k\leftarrow\frac{1}{n_k}\sum_{i=1}^n o_{ik} U^ty_i,\enspace
\hat{\Sigma}_k\leftarrow U^tC_kU,\enspace
\hat{\beta}_k\leftarrow\frac{trace(C_k)-\sum_{j=1}^d u_j^tC_ku_j}{p-d},\enspace
\end{equation*}
where $u_j$ is the $j$-th column of $U$ and $C_k=\frac{1}{n_k}\sum_{i=1}^n o_{ik} (y_i-m_k)(y_i-m_k)^t$ the empirical covariance matrix of the cluster $k$.

\subsection{Sparse version}
Yet, the use of latent space introduces dimension reduction and computation efficiency.
Nevertheless the back-projection from the latent space to the observation space can involve all the original features.
To do feature selection, the projection matrix $U$ has to be sparse.
\cite{bouveyron:hal-00685183} proposed 3 methods to enforce sparsity:
\begin{inparaenum}[1)]
 \item After a standard F-step, compute an sparse approximation of $U$ independently of the Fisher criterion,
 \item Compute the projection with a modified Fisher criterion with a $L_1$ penalty on $U$,
 \item Compute $U$ from the Fisher criterion using a penalized SVD algorithm. 
\end{inparaenum}

\section{Application to swimmer coordination}

The clustering is done in two steps:
\begin{inparaenum}[1)]
 \item A clustering on cycle data.
 Here an observation is just one swimming cycle. This clustering has two purposes, a) give a label to each cycle b) select which phase samples over the 100 are informative through sparsity.
 \item A clustering on trials.
 Each trial can be described now by a sequence of cycle labels learned at the first step. Features for this clustering consist in the transition matrix of the sequence with its diagonal put to zero.
\end{inparaenum}
The number of cluster is chosen by analysis of the Bayesian information criterion (BIC).

\begin{figure}
\begin{subfigure}[h]{0.5\textwidth}
\centering
\begin{tikzpicture}
\node[anchor=south west,inner sep=0] (fig) at (0,0) {\includegraphics[width=0.95\textwidth]{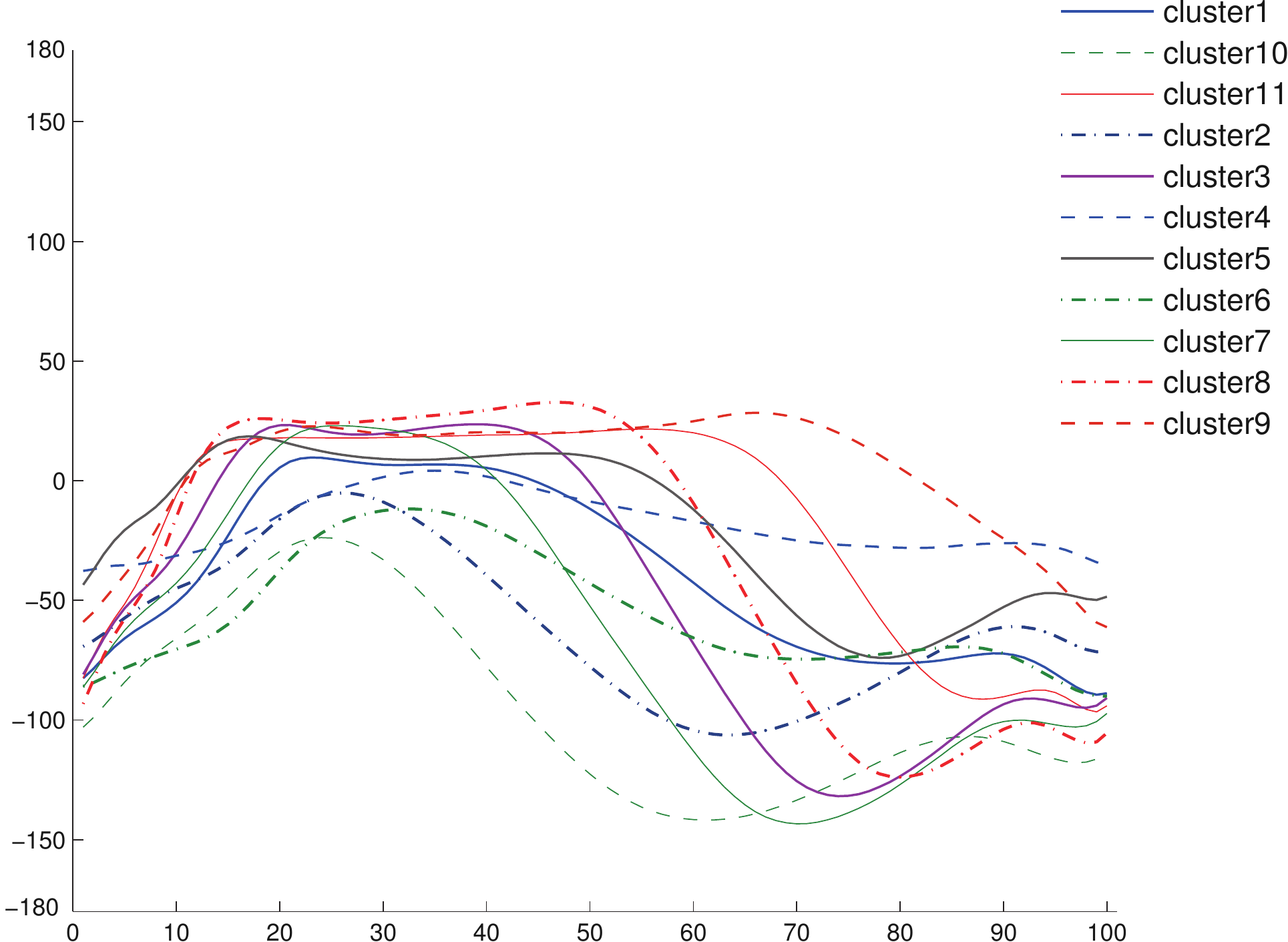}};
\path (fig.south)--+(0mm,-2mm) node {\tiny Percentage of cycle duration (\%)};
\path (fig.west)--+(-2mm,0mm) node[rotate=90] {\tiny Elbow-Knee relative phase (°)};
\end{tikzpicture}
\caption{a) Mean patterns of coordination for each cluster}
\label{fig:meanpat}
\end{subfigure}
~
\begin{subfigure}[h]{0.5\textwidth}
\centering
\begin{tikzpicture}
\node[anchor=south west,inner sep=0] (fig) at (0,0) {\includegraphics[width=0.95\textwidth]{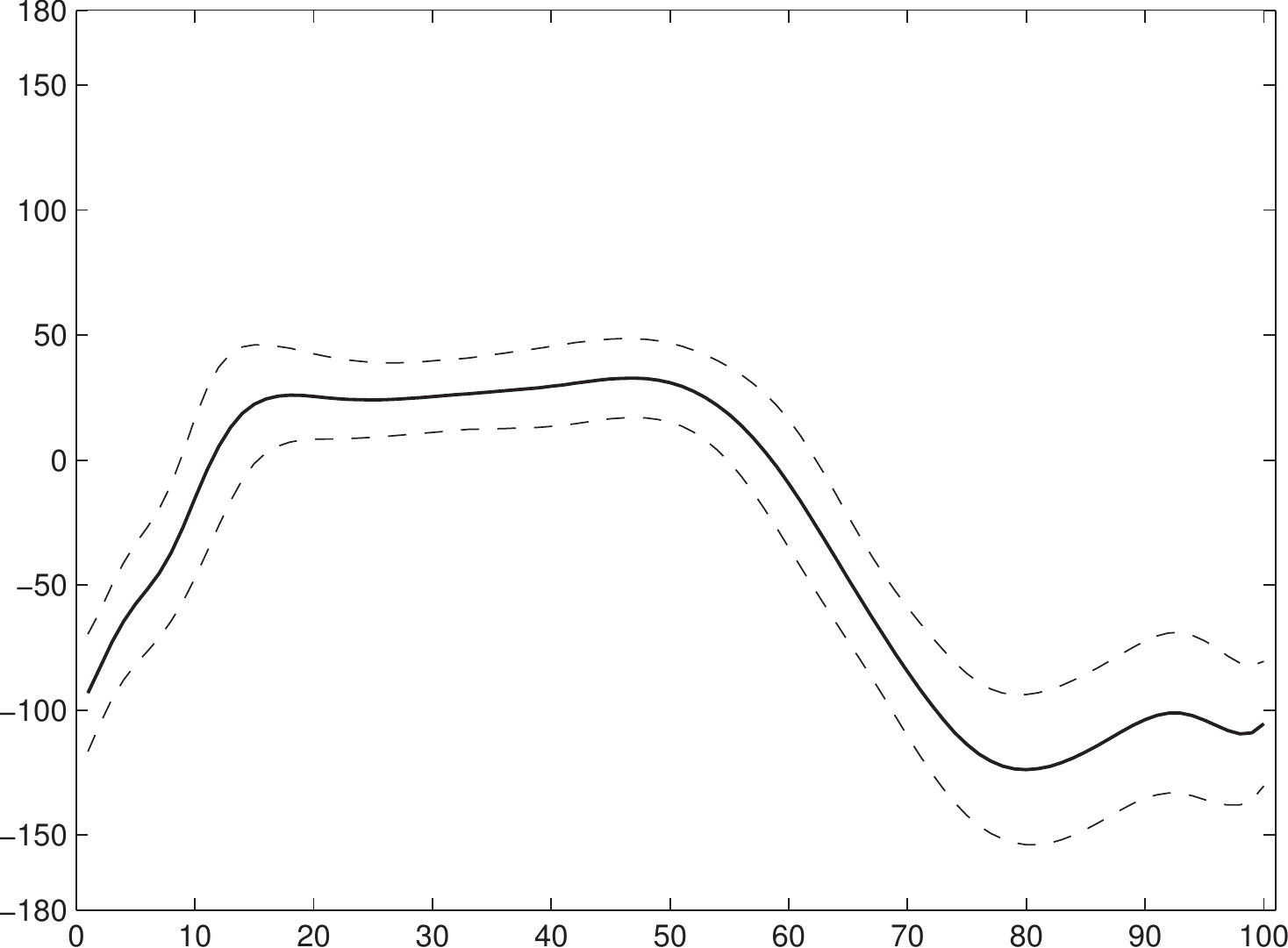}};
\path (fig.south)--+(0mm,-2mm) node {\tiny Percentage of cycle duration (\%)};
\path (fig.west)--+(-2mm,0mm) node[rotate=90] {\tiny Elbow-Knee relative phase (°)};
\end{tikzpicture}
\caption{b) Mean pattern for cluster 8 (black line), standard deviation (dotted line)}
\label{fig:c8}
\end{subfigure}

\begin{subfigure}[h]{\textwidth}
\centering
\begin{tikzpicture}
\node[anchor=south west,inner sep=0] (fig) at (0,0) {\includegraphics[width=0.8\textwidth]{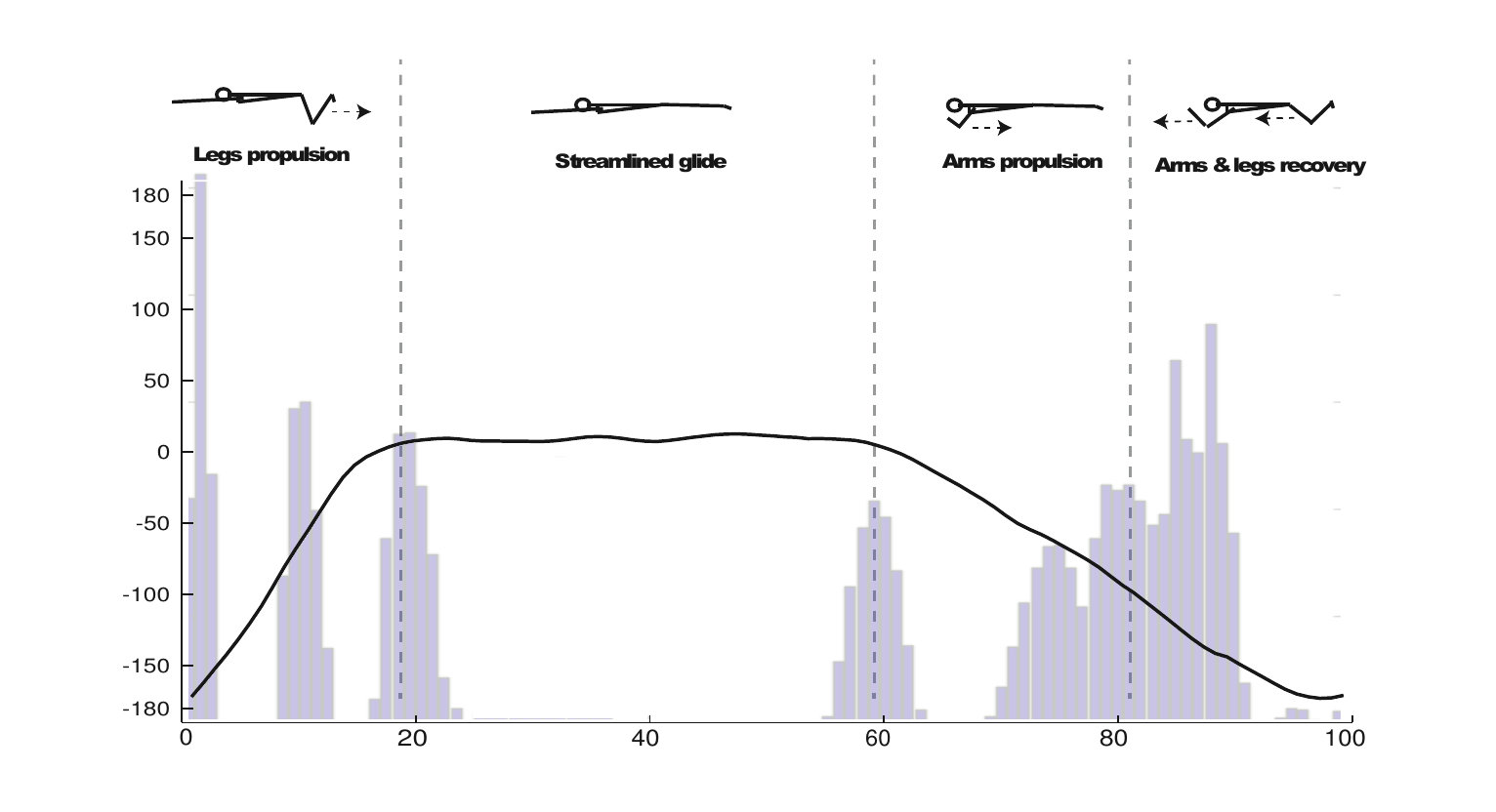}};
\path (fig.south)--+(0mm,+2mm) node {\tiny Percentage of cycle duration (\%)};
\path (fig.west)--+(+5mm,0mm) node[rotate=90] {\tiny Elbow-Knee relative phase (°)};
\end{tikzpicture}
\caption{c) A typical coordination and superimposed induced sparsity}
\label{fig:spartsity}
\end{subfigure}

\caption{First clustering level}
\end{figure}
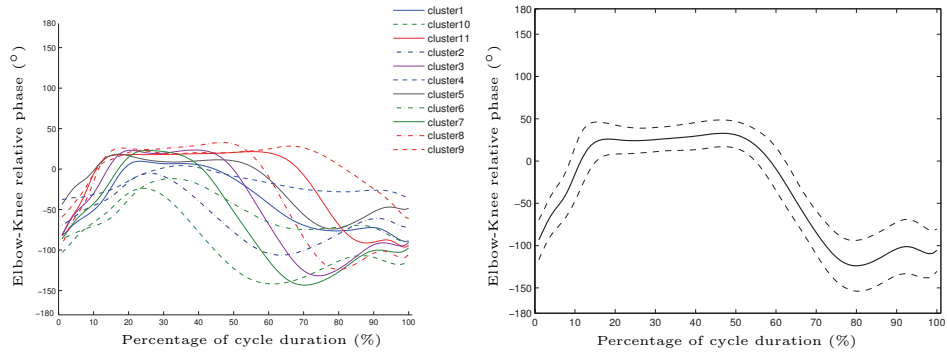
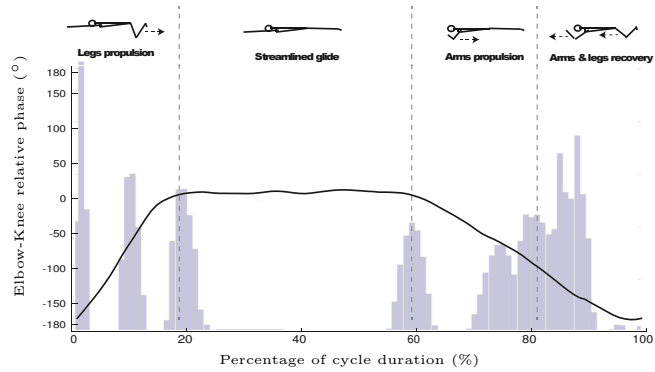

For the first clustering level, analysis of the BIC (Tab. \ref{tab:nbc:bic}) highlights the existence of 11 clusters within the whole set of data.
The mean coordination of these clusters are represented at Figure \ref{fig:meanpat}.

This result advocates for qualitative reorganizations of motor behavior during motor learning, as each learner visited between 9 and 11 different clusters during their sessions.
For instance, the mean and standard deviation of one cluster (n$^o$8) is presented in Figure \ref{fig:c8}.

In order to differentiate the effect of the different instructions on the learning process,
Table \ref{tab:distrib} shows the distribution of each emerging cluster across the different learning conditions.
Interestingly, the use of different additional instructions led to the exhibition of different preferred patterns of coordination.
For instance, the group who received an analogy exhibited preferably clusters 3, 7, 8 and 9, whereas clusters 2, 4 and 10 were inhibited.
In the meantime, the use of the prescriptive instruction preferably led to the use of cluster 5 and inhibited the use of clusters 2, 6 and 10.
This result is a key point of the experiment, validating the possibility of guiding the exploration during learning and by extension the result of the learning process
with using different types of instructions during the practice.

On Figure \ref{fig:spartsity}, we have superimposed a typical coordination curve and, in gray bars, the back-projection of latent space into observation space to see induced sparsity from the first level.
The height of a bar at a feature $i\in[1\ldots p]$ is proportional to $\sum_{j=1}^d |U_{ij}|$.
A null value shows that the corresponding feature is not involved in the projection to the latent space, i.e. it is not selected by the F-Step or it is squeezed by the sparsity;
therefore it can be considered not relevant to build the clusters.
Interestingly, only key points of the movement have high values, thus the Fisher-Em algorithm is able to select key points without any prior knowledge.

\begin{figure}
\centering
\begin{tikzpicture}
\node[anchor=south west,inner sep=0] (fig) at (0,0) {\includegraphics[width=0.6\textwidth]{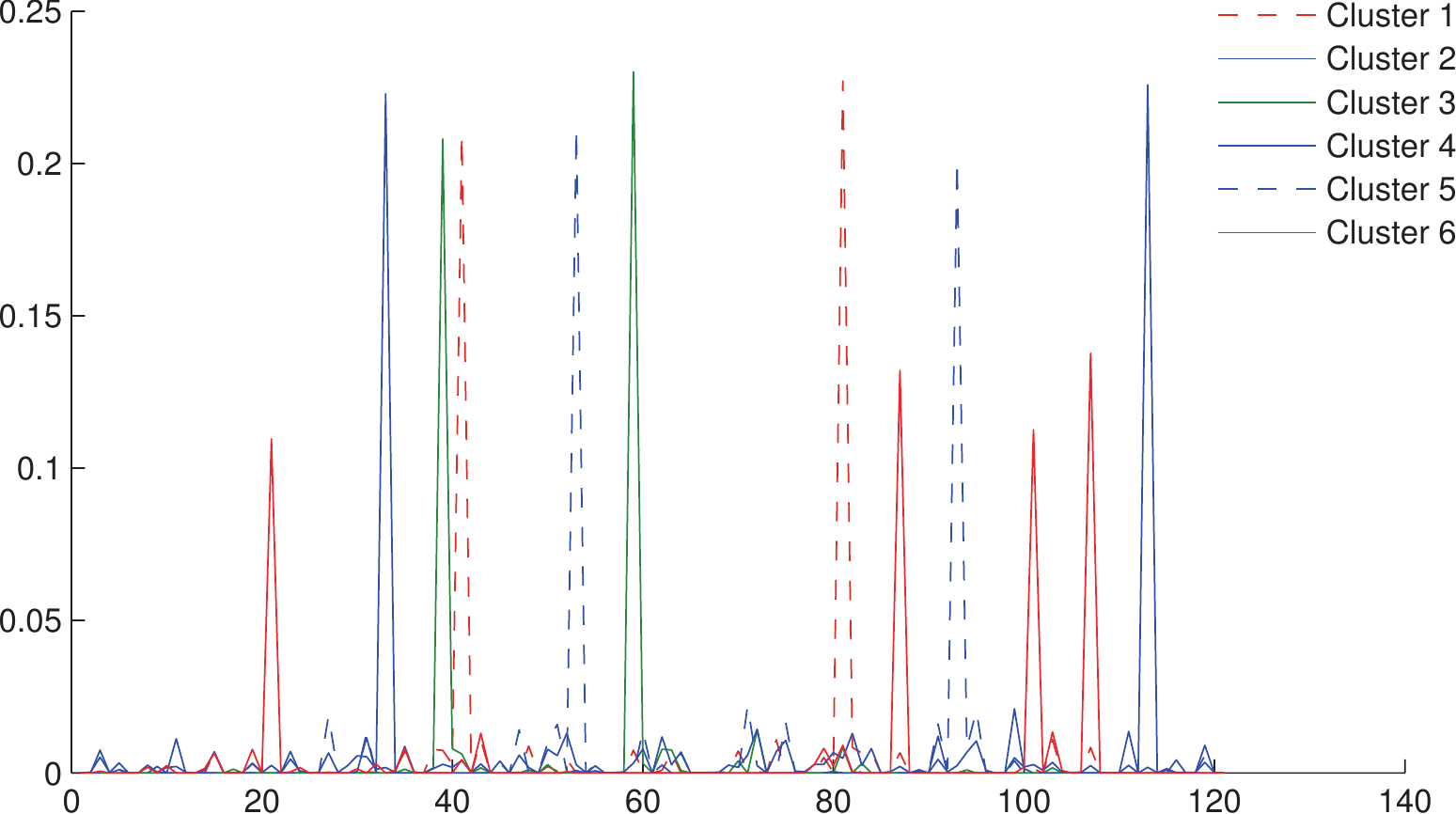}};
\path (fig.south)--+(0mm,-2mm) node {\tiny Transitions };
\path (fig.west)--+(-2mm,0mm) node[rotate=90] {\tiny Occurrences};
\end{tikzpicture}
\caption{Mean patterns of possible transitions within a trial for the 2nd level clustering, please note that there are $121=11 \times 11$ possible transitions as there is $11$ clusters at first level}
\label{fig:postrans}
\end{figure}

The second level of cluster analysis, based on the transition matrix during each trial showed the existence of six different clusters.
More specifically, Figure \ref{fig:postrans} highlights the preferred transitions exhibited by each emerging cluster.
Interestingly, the group who showed the highest number of preferred transition (i.e. cluster 6) was associated with the learning group that did not receive any instruction.
In that sense, this second level of cluster analysis allowed to highlight the use of temporary additional information during learning in order to modify the learning search strategy,
namely by impacting the preferred transitions. 

\section{Perspectives}
These preliminary experiments show that we can apply efficiently the Fisher-EM clustering on highly correlated features.
Interestingly, the induced sparsity corresponds to key points of the coordination phase.
Now, a qualitative work needs to be undertaken to qualify clusters of trials in term of learning condition and learning dynamics.

\begin{table}
\caption{Analysis of the BIC for the first level showing a plateau at 11 clusters}
\label{tab:nbc:bic}
\centering
\scriptsize
\begin{tabular}{|c||c|c|c|c|c|c|c|c|c|c|c|c|c|c|c|c|c|}
\hline
Number of clusters&2&3&4&5&6&7&8&9&10&11&12&13&14&15&16&17\\
\hline
BIC value ($\times 10^{7}$)&\tiny -1.23&\tiny -1.21&\tiny -1.18&\tiny -1.18&\tiny -1.15&\tiny -1.14&\tiny -1.13&\tiny -1.11&\tiny -1.08&\tiny -1.04&\tiny -1.05&\tiny -1.05&\tiny -1.07&\tiny -1.04&\tiny -1.04&\tiny -1.05\\
\hline
\end{tabular}
\end{table}

\begin{table}
\caption{Distribution (in \%) of each cluster according to learning conditions}
\label{tab:distrib}
\scriptsize
\centering
\begin{tabular}{|c|c|c|c|c|c||c|c|c|c|c|c|}
\hline
Cluster&Control&Analogy&Pacer&Prescription&Total&Cluster&Control&Analogy&Pacer&Prescription&Total\\
\hline
\hline
1&24.62&35.15&14.39&25.84&100&7&23.12&39.03&17.25&20.60&100\\
\hline
2&47.85&7.16&28.77&16.22&100&8&16.72&46.56&17.41&19.31&100\\
\hline
3&17.60&45.59&12.07&24.74&100&9&14.69&41.91&18.04&25.36&100\\
\hline
4&61.18&4.59&10.98&23.26&100&10&27.81&5.95&64.36&1.87&100\\
\hline
5&28.73&25.73&1.86&43.69&100&11&19.46&26.18&26.34&28.01&100\\
\hline
6&44.25&16.70&23.95&15.09&100&&&&&&\\
\hline

\end{tabular}
\end{table}

%
%
\FloatBarrier
\newpage
\bibliographystyle{splncs}
\bibliography{ecml2013komarherault}

\end{document}